\PassOptionsToPackage{table}{xcolor}
\documentclass[sigconf]{acmart}
\usepackage{booktabs}
\usepackage{graphicx}
\usepackage{makecell}
\usepackage{pifont}
\usepackage{multirow}
\usepackage{subcaption}
\definecolor{HeaderGray}{gray}{0.9}
\definecolor{myExtremelyLightBlue}{RGB}{235,245,255}
\definecolor{markgreen}{RGB}{0,128,70}
\definecolor{markred}{RGB}{190,40,40}

\usepackage{xcolor}
\usepackage{graphicx}
\usepackage{wrapfig}
\usepackage[most]{tcolorbox}

\definecolor{correctgreen}{RGB}{0,150,80}

\newtcolorbox{ReasoningBox}[2][]{
    enhanced,
    breakable, 
    colback=white,
    colframe=black,
    coltitle=white,
    fonttitle=\bfseries,
    title={#2},
    sharp corners,
    boxrule=1pt,
    attach boxed title to top left={xshift=0mm, yshift=0mm},
    boxed title style={
        sharp corners, 
        colback=black, 
        colframe=black, 
        boxrule=1pt,
        right=2mm, left=2mm
    },
    top=1.5em,
    #1
}

\newcommand{\ours}{\textsc{CLIR-Bench}}

\newcommand{\cmark}{\textcolor{markgreen}{\ding{51}}}
\newcommand{\xmark}{\textcolor{markred}{\ding{55}}}

\newcommand{\eat}[1]{}

\definecolor{myExtremelyLightGreen}{RGB}{232, 245, 233}
\definecolor{tablerowcolor}{RGB}{242,242,242}
\definecolor{tablerowcolor1}{RGB}{242,242,242}
\AtBeginDocument{%
  }

\setcopyright{acmlicensed}
\copyrightyear{2018}
\acmYear{2018}
\acmDOI{XXXXXXX.XXXXXXX}
\acmConference[Conference acronym 'XX]{Make sure to enter the correct
  conference title from your rights confirmation email}{June 03--05,
  2018}{Woodstock, NY}
\acmISBN{978-1-4503-XXXX-X/2018/06}

\begin{document}

\title{\ours: Benchmarking Multimodal Question Answering over Irregular Clinical Time Series}

\author{Frank Nie}
\authornote{Equal Contribution.}
\affiliation{%
  \institution{Shandong University}
  \country{China}
}

\author{Ethan B. Liu}
\authornotemark[1]
\authornote{Corresponding author.}
\affiliation{%
  \institution{Shandong University}
  \country{China}
}

\author{Yuan Zhu}
\authornotemark[1]
\affiliation{%
  \institution{Shandong University}
  \country{China}
}
\email{zhu948982@gmail.com}

\author{Loe Yan}
\affiliation{%
  \institution{Shandong University}
  \country{China}
}

\author{Wei Fan}
\affiliation{%
  \institution{University of Auckland}
  \country{New Zealand}
}

\author{Jindong Han}
\authornotemark[2]
\affiliation{%
  \institution{Shandong University}
  \country{China}
}
\email{jindong.han@sdu.edu.cn}

\renewcommand{\shortauthors}{First Author et al.}

\renewcommand{\shortauthors}{Trovato et al.}


\begin{abstract}
Clinical time series are central to patient monitoring, risk assessment, and clinical decision support. However, they are often sparse, irregularly sampled, and asynchronous, making it difficult for models to identify the temporal evidence required for clinical Question Answering (QA). Existing benchmarks primarily focus on regularly sampled time-series QA or medical QA over static data, and therefore rarely assess whether models can faithfully ground their answers in irregular temporal observations. To fill this gap, we introduce \ours{}, a benchmark for irregular clinical time series QA constructed from de-identified ICU records through a principled four-stage pipeline. \ours{} contains 6,600 QA instances spanning 11 clinical variables, organized into four capability dimensions and 11 tasks. Each question is linked to explicit temporal evidence and task-specific answer derivation rules, enabling evaluation of both answer accuracy and evidence use. Experiments show that existing generalist models struggle to retrieve and reason over sparse clinical evidence, highlighting the need for stronger irregular time-series reasoning methods. Our code and data are available at \hyperlink{}{https://huggingface.co/datasets/winall/CLIR-Bench}.
\end{abstract}



\maketitle

\section{Introduction}
\label{sec:benchmark-construction}
\begin{figure*}[t]
  \centering
  \includegraphics[width=0.98\textwidth]{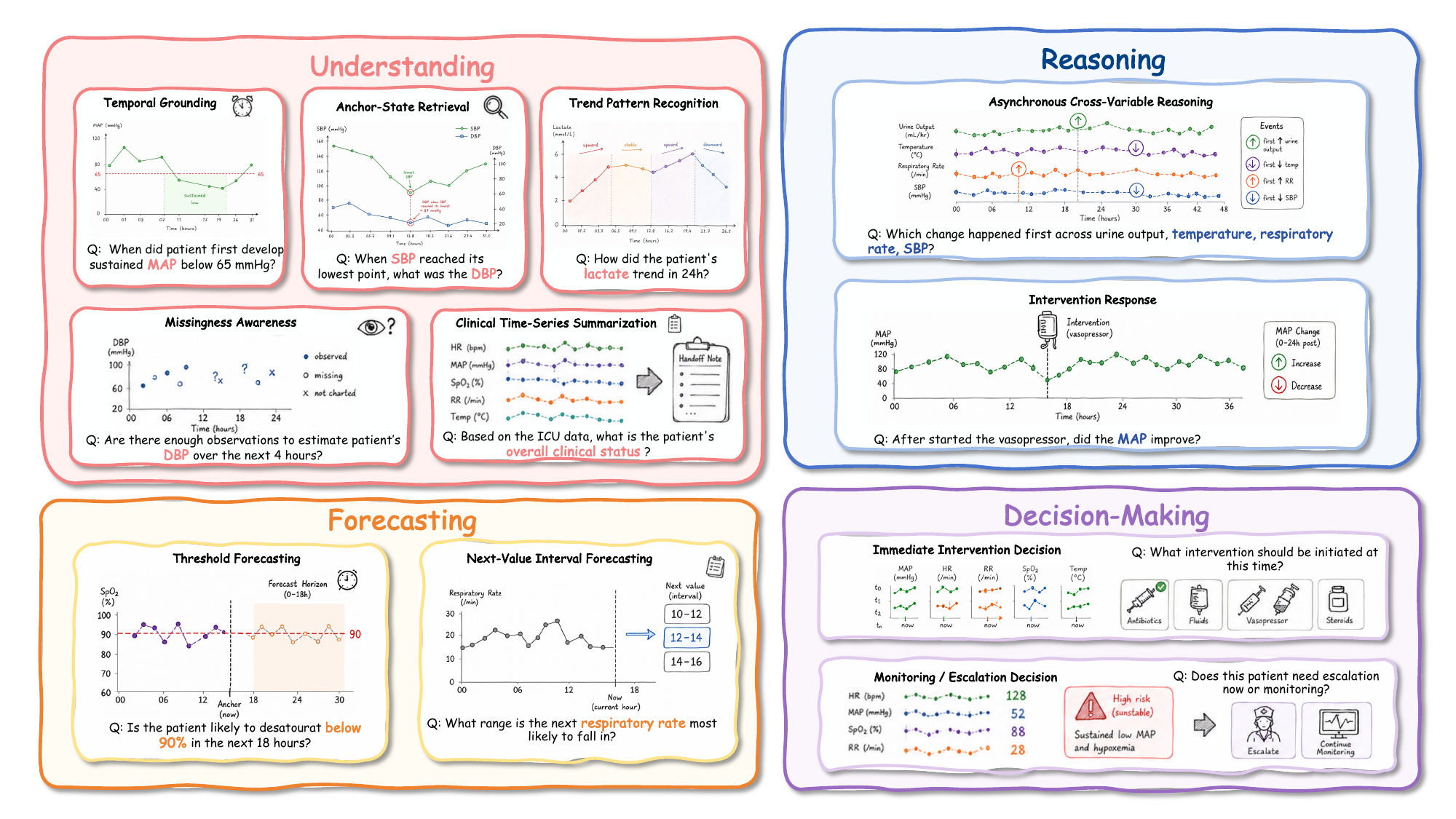}
  \caption{Task categories of the \ours{}.}
  \label{fig:clirbench-case}
\end{figure*}
Clinical time series play a pivotal role in patient monitoring, risk assessment, and clinical decision support~\citep{li2025miramedicaltimeseries,zheng2026rethinkinglargelanguagemodels}. In intensive care units (ICU), clinicians continuously interpret vital signs, laboratory measurements, and treatment events to understand how a patient’s condition changes over time~\citep{PhysioNet-mimiciv-3.1,bui2024benchmarkingmimicivirregularspare}. Unlike standard time series, however, clinical time series are often sparse, irregularly sampled, and asynchronous across variables~\citep{chung2026milmlargelanguagemodels}. Some measurements (e.g., heart rate) may be recorded every few minutes, while others, such as lactate, may appear only a few times over many hours. As a result, the information needed to answer a clinical question is often limited to a small number of observations scattered across a long patient record. The key challenge is therefore to identify the relevant observations, align events recorded at different times, and use them as evidence for a reliable answer~\citep{zhang2024largelanguagemodelstime}.

Large Language Models (LLMs) offer a natural-language interface for querying such complex patient records~\citep{Omiye_2024,fleming2023medaligncliniciangenerateddatasetinstruction,du2026medhorizonlongcontextmedicalvideo}. Instead of requiring clinicians to manually inspect long and irregular patient trajectories, LLMs could help interpret temporal patterns and answer questions about patient states and treatment responses~\citep{chan2024medtsllmleveragingllmsmultimodal,cui2025timertemporalinstructionmodeling}. However, this capability also raises a critical evaluation challenge: does a model answer correctly because it has used the relevant patient-specific observations, or because it relies on general clinical knowledge or other shortcuts~\citep{kruse2025largelanguagemodelstemporal,pandit2025medhallucomprehensivebenchmarkdetecting}? In safety-sensitive settings, an answer that is not supported by the correct temporal evidence may be unreliable even when it happens to be correct.

Existing benchmarks do not fully address this need. Traditional time-series benchmarks primarily focus on forecasting, classification, or anomaly detection, where outputs are typically fixed numerical values or predefined labels~\citep{qiu2025tfbcomprehensivefairbenchmarking,qiu2025tabunifiedbenchmarkingtime}. Recent time-series Question-Answering (QA) benchmarks extend evaluation to natural-language questions, but many of them are built on regular, aligned, or well-structured sequences~\citep{kong2025timemqatimeseriesmultitask,jing2026tsaqatimeseriesanalysis}. In parallel, medical QA benchmarks have made strong progress in evaluating reasoning over clinical text, medical images, structured knowledge, and static patient information~\citep{yan2025largelanguagemodelbenchmarks,hu2024omnimedvqanewlargescalecomprehensive}. Despite fruitful progress, existing benchmarks rarely assess whether models can locate, integrate, and faithfully use temporal evidence from sparse and asynchronous time series for answering clinical meaningful questions.

To address this gap, we introduce \ours{}, a benchmark for multimodal question answering over irregular clinical time series. \ours{} is developed through a principled four-stage pipeline: (1) time series data extraction from de-identified ICU records, (2) task instantiation to encode analytical goals in executable and verifiable form, (3) QA pairs generation with explicit temporal evidence, and (4) human verification to ensure the quality of generated samples. Each QA instance is linked to timestamp-level evidence and a task-specific rule for deriving the correct answer. This design makes the benchmark evidence-auditable, where the supporting evidence can be isolated, removed, or edited for controlled evaluation.  Overall, \ours{} contains 6,600 QA instances, covering 11 clinical variables together with intervention signals. The benchmark is organized around four core capability dimensions: temporal understanding, reasoning, forecasting, and decision-making, yielding 11 clinically motivated tasks in total.

Our contributions are threefold:
\begin{itemize}
  \item We introduce \ours{}, an irregular clinical time-series question-answering benchmark. It contains 6,600 QA instances grounded in ICU trajectories, covering 11 clinical variables and 11 clinically motivated tasks.
  \item We propose an evidence-auditable benchmark design. Each QA instance is associated to explicit temporal evidence, enabling evaluation beyond answer accuracy, including evidence faithfulness, evidence reliance, and sensitivity to answer-determining observations.
  \item We provide a diagnostic evaluation of current models. Experiments show that existing LLMs struggle to reliably retrieve and reason over sparse temporal evidence, highlighting the need for stronger evidence selection methods and native irregular time-series reasoning models.
\end{itemize}

\section{Related Work}

\begin{table*}[t]
\centering
\caption{Comparison with representative time-series QA benchmarks.}
\label{tab:benchmark-comparison}
\begingroup
\footnotesize
\setlength{\tabcolsep}{10pt}
\setlength{\aboverulesep}{0.4ex}
\setlength{\belowrulesep}{0.4ex}
\resizebox{0.92\textwidth}{!}{%
\begin{tabular}{lcccccccccc}
\toprule
\multirow{2}{*}{\textbf{Benchmark}}
& \multicolumn{6}{c}{\textbf{Scope \& Coverage}}
& \multicolumn{4}{c}{\textbf{Temporal Capabilities}} \\
\cmidrule(lr){2-7} \cmidrule(lr){8-11}
& \textbf{Task}
& \textbf{Mul.}
& \textbf{QA}
& \textbf{Irregular}
& \textbf{QA}
& \textbf{Missing-Data}
& \textbf{Underst.}
& \textbf{Reason.}
& \textbf{Forecast.}
& \textbf{Decision.} \\
\midrule
GIFT-Eval~\cite{aksu2024giftevalbenchmarkgeneraltime}
& $1$ & \xmark & \xmark & \xmark & -- & \xmark & \xmark & \xmark & \cmark & \xmark \\
TimeSeriesExam~\cite{cai2024timeseriesexamtimeseriesunderstanding}
& $5$ & \cmark & \cmark & \xmark & $0.7$K & \xmark & \cmark & \cmark & \xmark & \xmark \\
EngineMT-QA~\cite{wang2025itformerbridgingtimeseries}
& $4$ & \cmark & \cmark & \xmark & $11$K & \xmark & \cmark & \cmark & \xmark & \xmark \\
Time-MQA~\cite{kong2025timemqatimeseriesmultitask}
& $5$ & \cmark & \cmark & \xmark & $200$K & \xmark & \cmark & \cmark & \cmark & \xmark \\
TSRBenchmark~\cite{yu2026tsrbenchcomprehensivemultitaskmultimodal}
& $15$ & \cmark & \cmark & \xmark & $4.1$K & \xmark & \cmark & \cmark & \cmark & \cmark \\
TSAQA~\cite{jing2026tsaqatimeseriesanalysis}
& $6$ & \cmark & \cmark & \xmark & $210$K & \xmark & \cmark & \cmark & \cmark & \xmark \\
MMTS-BENCH~\cite{yin2026mmtsbenchcomprehensivebenchmarktime}
& $19$ & \cmark & \cmark & \xmark & $2.4$K & \xmark & \cmark & \cmark & \xmark & \xmark \\
ARFBench~\cite{xie2026arfbenchbenchmarkingtimeseries}
& $8$ & \cmark & \cmark & \xmark & $0.75$K & \xmark & \cmark & \cmark & \xmark & \cmark \\
\midrule
\rowcolor{myExtremelyLightGreen}
\textbf{\ours{}}
& $\textbf{11}$ & \cmark & \cmark & \cmark & $\textbf{6.6}$K & \cmark & \cmark & \cmark & \cmark & \cmark \\
\bottomrule
\end{tabular}%
}
\endgroup
\end{table*}

\textbf{Time series analysis.} Time series analysis has traditionally centered on forecasting, classification, anomaly detection, and imputation~\citep{Lim_2021,Ismail_Fawaz_2019,blázquezgarcía2020reviewoutlieranomalydetectiontime,Liu_2025,nie2026stoprewardinghallucinatedsteps,liu2026divide}. To support systematic evaluation of progress on these tasks, both classical statistical methods and modern learning-based models have been assessed using a diverse set of benchmark suites. For example, the UCR~\citep{dau2019ucrtimeseriesarchive} and UEA~\citep{bagnall2018ueamultivariatetimeseries} archives support standardized evaluation for univariate and multivariate time series classification, while TFB~\citep{qiu2025tfbcomprehensivefairbenchmarking} and GIFT-Eval~\citep{aksu2024giftevalbenchmarkgeneraltime} provide broader evaluation settings for forecasting models across datasets, domains, and prediction horizons. These benchmarks have enabled reproducible comparisons and accelerated progress in time-series modeling. Nevertheless, they primarily assess numerical prediction or assignment to predefined task labels. Consequently, they do not directly evaluate whether a model can interpret, integrate, and reason about temporal observations to answer open-ended natural-language questions.

\textbf{Time series QA benchmarks.} Recent time series QA benchmarks extend evaluation beyond fixed-output prediction by formulating temporal analysis tasks as natural-language questions~\citep{jiang2024empoweringtimeseriesanalysis,zhang2024largelanguagemodelstime}. TimeSeriesExam~\citep{cai2024timeseriesexamtimeseriesunderstanding}, for example, introduces an exam-style benchmark for assessing general time series understanding, whereas ChatTS~\citep{Xie_2025} investigates the alignment of time series representations with language for efficient interaction. Time-MQA~\citep{kong2025timemqatimeseriesmultitask} constructs a large-scale multi-task QA dataset covering forecasting, imputation, anomaly detection, classification, and reasoning. Other benchmarks extend time series QA to multimodal or domain-specific settings. MTBench~\citep{chen2026mtbenchmultimodaltimeseries} combines financial and weather time series with textual reports, while EngineMT-QA~\citep{wang2025itformerbridgingtimeseries} aligns industrial multivariate sensor data with natural language questions. TSAQA~\citep{jing2026tsaqatimeseriesanalysis} offers a unified evaluation framework spanning multiple domains and task formats, and MMTS-BENCH~\citep{yin2026mmtsbenchcomprehensivebenchmarktime} further evaluates multimodal reasoning and cross-modal alignment. ARFBench~\citep{xie2026arfbenchbenchmarkingtimeseries}, in contrast, focuses specifically on anomaly reasoning in operational telemetry and incident-response scenarios. Collectively, these benchmarks greatly broaden time series evaluation. However, existing time series QA benchmarks largely assume regularly sampled and well-structured observations, while real-world clinical time series are typically sparse and asynchronous~\citep{tipirneni2022selfsupervisedtransformersparseirregularly, shukla2021multitimeattentionnetworksirregularly}. To fill this gap, we provide a focused testbed for evaluating whether current LLMs can interpret and integrate sparse temporal evidence to answer clinically meaningful questions.

\section{\ours{}}

\subsection{Problem Formulation}
We define question answering over irregular clinical time series as follows. Given a patient-specific ICU trajectory $\mathcal{X}=\{(t_i,v_i,x_i,c_i)\}_{i=1}^{L}$, where $L$ denotes the number of observed entries and each tuple consists of a timestamp $t_i$, clinical variable $v_i$, measured value $x_i$, and optional intervention context $c_i$, the goal is to answer a natural language question $q$ by reasoning over the patient’s irregular, asynchronous, and sparse clinical record. Formally, the task can be written as $\mathcal{F}: (q,\mathcal{X}) \mapsto \mathcal{A}$, where $\mathcal{A}$ is the generated answer and $\mathcal{F}$ denotes a model that interprets the question $q$ and analyzes the irregularly sampled time series $\mathcal{X}$.



\subsection{Dataset and Task Design}
\ours{} contains 6,600 QA instances spanning 11 clinical variables and intervention signals. It evaluates generalist models across four core capability dimensions for irregular clinical time-series QA: understanding, reasoning, forecasting, and decision-making. As shown in Figure~\ref{fig:clirbench-case}, these four dimensions are further divided into 11 tasks, providing a structured assessment of model performance under realistic temporal irregularity.

\subsubsection{Data Curation}
\ours{} is built from MIMIC-IV~\citep{PhysioNet-mimiciv-3.1}, a large-scale critical care database containing de-identified ICU records. We focus on patient-level time-series trajectories because ICU monitoring is naturally sparse, asynchronous, and shaped by clinical interventions. The curated cohort contains 500 ICU stays. For each ICU stay, we extract a set of commonly monitored vital signs, laboratory measurements, urine output, and intervention-related events. The vital signs include heart rate, mean arterial pressure (MAP), systolic blood pressure (SBP), diastolic blood pressure (DBP), peripheral oxygen saturation (SpO$_2$), respiratory rate, and temperature. The laboratory variables include lactate, creatinine, and white blood cell count. Additionally, each record includes elapsed ICU time, observation masks, and time-gap features indicating the time since the previous observation of the same variable. Please refer Appendix~\ref{apx:data fields} for detailed data description.

\begin{figure}[t]
  \centering
  \includegraphics[width=0.42\textwidth]{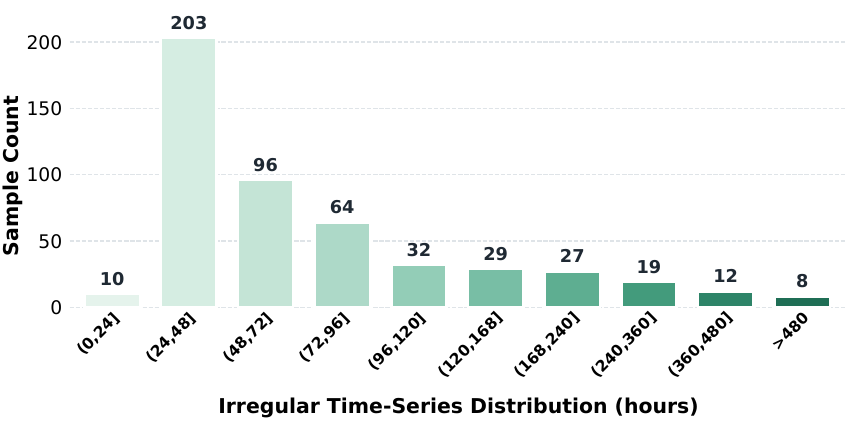}
  \caption{Duration distribution of time-series samples.}
\label{fig:clirbench-duration}
\end{figure}

\begin{figure}[t]
  \centering
  \includegraphics[width=0.4\textwidth]{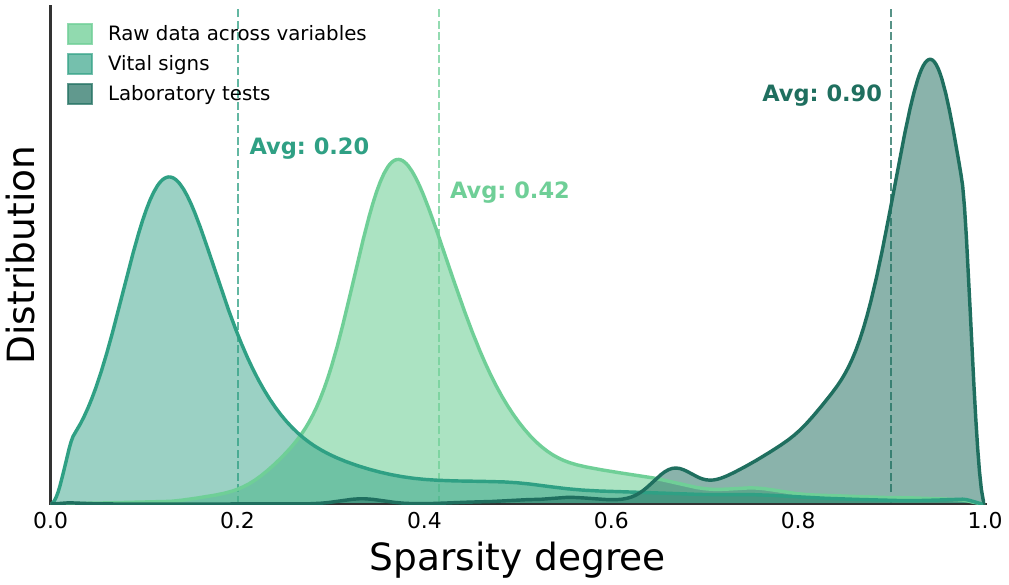}
  \caption{Observation sparsity distribution in \ours{}. 
The sparsity degree denotes the fraction of missing variable-time cells in QA instance.}
\label{fig:clirbench-sparsity}
\end{figure}

We summarize the temporal characteristics of the curated dataset below. Figure~\ref{fig:clirbench-duration} shows the distribution of ICU trajectory durations. Many trajectories span multi-day ICU stays, with a large portion lasting between 24 and 96 hours and a long tail extending beyond 240 or even 480 hours. This avoids restricting the benchmark to short, regular windows and instead reflects the long-horizon nature of ICU monitoring. Figure~\ref{fig:clirbench-sparsity} further highlights the irregularity of the data. Vital signs are relatively dense, with an average sparsity degree of 0.20, whereas laboratory tests are much sparser, with an average sparsity degree of 0.90. These properties distinguish \ours{} from time-series QA settings that assume regular sampling on a shared temporal grid.

\subsubsection{Task Categories}
\ours{} covers 11 subtasks grouped into four capability dimensions: temporal understanding, reasoning, forecasting, and decision-making.

\textbf{Temporal understanding} evaluates whether a model can interpret irregular clinical observations, which includes five subtasks. \emph{Temporal Grounding} (TG) asks the model to locate relevant time points or intervals, such as the first occurrence of a sustained abnormal state. \emph{Anchor-State Retrieval} (ASR) queries the value of one variable at a clinically defined anchor time, such as the value of DBP when SBP reaches its lowest point. \emph{Trend Pattern Recognition} (TPR) requires the model to identify the temporal pattern of a variable within a specified window. \emph{Missingness Awareness} (MA) evaluates whether the available observations are sufficient to support a reliable answer. \emph{Clinical Time-Series Summarization} (TSS) asks the model to summarize the overall clinical status reflected by the patient trajectory.

\textbf{Temporal reasoning} evaluates whether a model can integrate information across variables and relate changes in patient state to clinical interventions. \emph{Asynchronous Cross-Variable Reasoning} (CVR) asks the model to compare the timing, direction, or relationship of changes across different clinical variables. \emph{Intervention Response} (IR) evaluates whether the model can characterize how a patient’s measurements change after a clinical intervention.

\textbf{Temporal forecasting} evaluates whether a model can infer future clinical states from the observed irregular history. \emph{Threshold Forecasting} (TF) asks whether a variable will cross a clinically meaningful threshold within a future time window. \emph{Next-Value Interval Forecasting} (NIF) asks the model to predict the interval into which the next observed measurement is likely to fall. For both tasks, a cutoff time separates the model-visible history from the future label-generation window, ensuring that future observations are not leaked into the input.

\textbf{Temporal decision-making} evaluates whether a model can connect irregular temporal evidence with retrospective clinical decision labels in a controlled benchmark setting. \emph{Immediate Intervention Decision} (IID) asks which intervention category is most consistent with current clinical state and recent history. \emph{Monitoring/Escalation Decision} (MED) asks whether the patient should remain under standard monitoring or be assigned closer monitoring based on recent measurements and risk signals.

\begin{figure*}[t]
  \centering
  \includegraphics[width=0.90\textwidth]{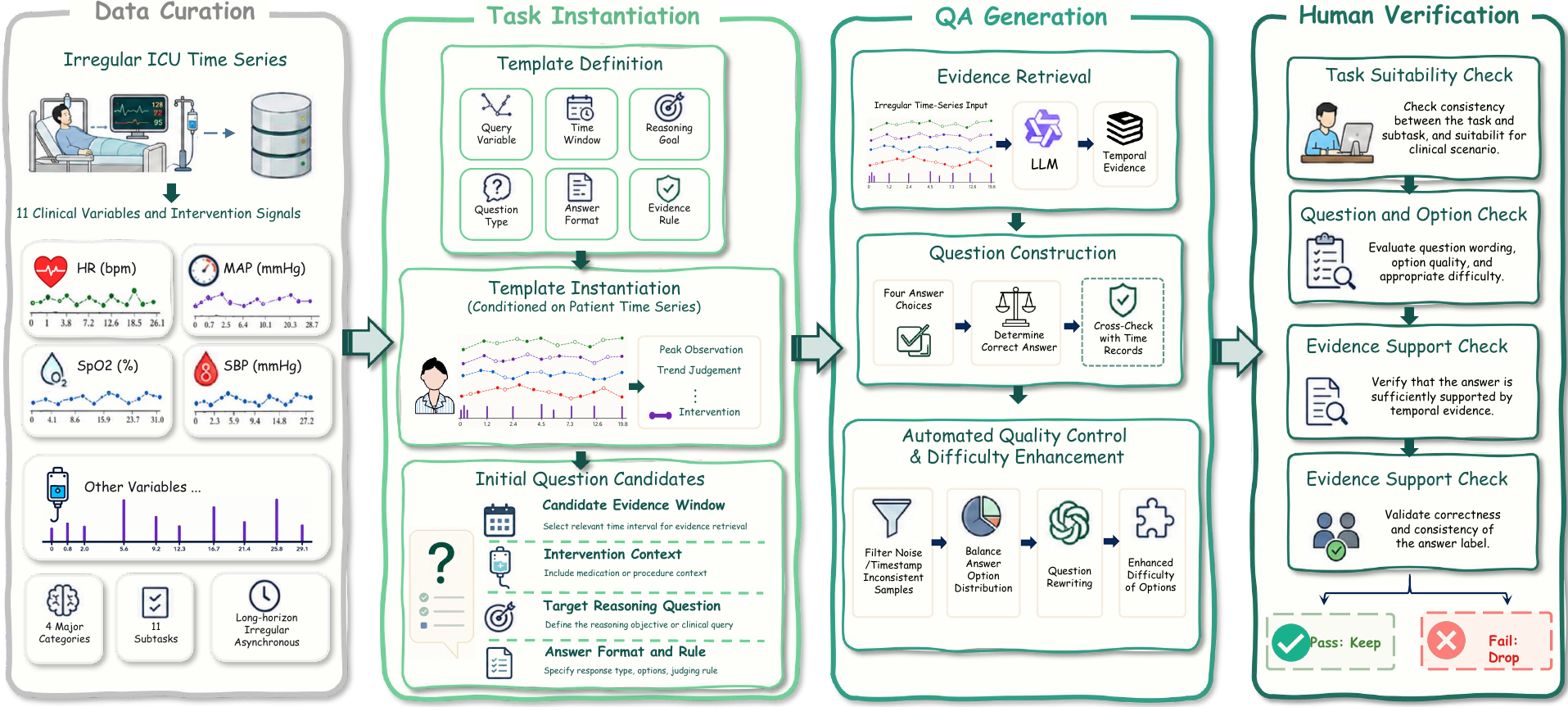}
  \caption{Overview of the construction pipeline. Starting from irregular time-series records, we first instantiate task-specific schemas, then generate candidate QA instances with explicit temporal evidence, and finally conduct human verification to ensure the quality of generated samples.}
  \label{fig:clirbench-pipeline}
\end{figure*}

\subsection{Construction Pipeline}
As shown in Figure~\ref{fig:clirbench-pipeline}, \ours{} is constructed through an evidence-grounded pipeline that converts irregular time series into clinically meaningful QA instances. Starting from the curated patient records described in Section 3.2, the pipeline consists of three stages: task instantiation, QA generation, and human verification.

\subsubsection{Task Instantiation}
First of all, we define a set of task-specific schemas for the 11 tasks. Each schema specifies the queried clinical variables, model-visible time window, reasoning goal, answer format, and evidence rule used to derive the gold label. For each patient trajectory, we instantiate these schemas by selecting valid evidence windows, identifying relevant intervention contexts when needed, and mapping timestamped records to target questions. This step produces initial question candidates with structured evidence requirements. Each candidate is associated with a task type, a source patient stay, a visible time span, a candidate evidence window, and a deterministic answer rule.

\subsubsection{QA Generation}
For each candidate, we generate a multiple-choice QA instance from the irregular time-series input. A language model is used to retrieve candidate temporal evidence, draft the question, and generate answer options. The correct answer, however, is determined by the task-specific evidence rule and cross-checked against the original timestamped records rather than relying on the generated text alone. We then apply automatic quality control to remove invalid or unreliable samples, including cases with inconsistent timestamps, duplicated options, or multiple correct choices. For forecasting tasks, we additionally verify that the prompt contains only pre-cutoff history and that the answer is derived only from the future label window. After filtering, answer options are shuffled to balance answer positions, distractors are strengthened with clinically plausible alternatives, and questions are rewritten into concise natural language while preserving the original evidence rule and gold label.

\subsubsection{Human Verification}
The final stage uses human verification as an audit layer. Each QA instance is checked for task suitability, question and option quality, evidence support, and answer-label consistency. Reviewers verify that the item matches the intended task, that the question is understandable, that the options are mutually exclusive, and that the gold answer is supported by the corresponding timestamped observations or intervention events. Samples that pass all checks are included in \ours{}, while samples with unclear wording, insufficient evidence, ambiguous answers, or inconsistent labels are removed or returned for revision. More details about the pipeline are provided in Appendix~\ref{apx:construction_audit}.

\section{Experiments}
\label{sec:experiments}


\subsection{Experimental Setup}
\label{subsec:experimental-setup}
\textbf{Models.} We evaluate a diverse set of baseline models, including (1) closed-source LLMs: Gemini-2.5-flash and GPT-5.4 mini, (2) open-source LLMs: DeepSeek-V4-flash, KiMi-2.6, Gemma3-27B-it, Gemma4-12B-it, Qwen3.5-4B~\citep{qwen35blog}, Qwen3.5-9B~\citep{qwen35blog}, Qwen3.6-27B~\citep{qwen36plus}, and (3) time-series LLMs: TimeOmni-1~\citep{guan2026timeomni1incentivizingcomplexreasoning}, TS-Reasoner~\citep{yu2025tsreasoneraligningtimeseries}, ITFormer~\citep{wang2025itformerbridgingtimeseries}, 
ChatTS~\citep{Xie_2025},
AutoTime~\citep{liu2024autotimesautoregressivetimeseries}, TimeLLM~\citep{jin2024timellmtimeseriesforecasting}. For AutoTime and TimeLLM, we follow the adaptation setting used by ITFormer~\citep{wang2025itformerbridgingtimeseries} so that these models can process time-series inputs under a unified configuration.

\textbf{Input settings.} We evaluate models under four input settings: (1) \emph{Full-TS} provides the question, answer options, and the serialized irregular time series, (2) \emph{QA-only} removes time series and keeps only the question and options, (3) \emph{Evidence-only} keeps only the gold supporting observations, intervention events, or evidence windows, (4) \emph{Evidence-removed} deletes the gold evidence while keeping the remaining time series context.

\textbf{Evaluation metrics.} For the main benchmark evaluation, we use answer accuracy as the primary metric. Since all tasks are formulated as four-way multiple-choice QA, the random baseline is 25\%. We report task-level accuracy for each subtask and macro-average accuracy across all 11 subtasks as the overall score. For other diagnostic analysis, we introduce the corresponding metrics in the relevant sections.

\begin{table*}[t]
\centering
\setlength{\tabcolsep}{3.0pt}
\renewcommand{\arraystretch}{1.15}
\caption{Model performance on \ours{}. Task abbreviations follow Section 3.2.2. Overall denotes the macro-average across all 11 tasks, and bold marks the best result in each column.}
\label{tab:model_task_accuracy}
\resizebox{0.9\textwidth}{!}{%
\begin{tabular}{lcccccccccccc}
\toprule
\textbf{Model}
& \multicolumn{5}{c}{\textbf{Understanding}}
& \multicolumn{2}{c}{\textbf{Forecasting}}
& \multicolumn{2}{c}{\textbf{Reasoning}}
& \multicolumn{2}{c}{\textbf{Decision-Making}}
& \textbf{Overall} \\
\cmidrule(lr){2-6}\cmidrule(lr){7-8}\cmidrule(lr){9-10}\cmidrule(lr){11-12}\cmidrule(lr){13-13}
& \textbf{TG}
& \textbf{ASR}
& \textbf{TPR}
& \textbf{MA}
& \textbf{TSS}
& \textbf{TF}
& \textbf{NIF}
& \textbf{CVR}
& \textbf{IR}
& \textbf{IID}
& \textbf{MED}
& \textbf{Avg.} \\
\midrule
\rowcolor{tablerowcolor1}
\multicolumn{13}{l}{$\blacktriangledown$ \emph{Closed-source LLMs}} \\
Gemini-2.5-flash & 41.67 & 43.33 & 25.00 & 23.33 & 91.67 & 20.00 & 10.00 & 43.33 & 60.00 & 60.00 & 25.00 & 40.30 \\
GPT-5.4 mini & \textbf{58.33} & \textbf{63.33} & \textbf{35.00} & 45.00 & \textbf{93.33} & 33.33 & 6.67 & \textbf{63.33} & 51.67 & 76.67 & 25.00 & \textbf{50.15} \\
\midrule
\rowcolor{tablerowcolor}
\multicolumn{13}{l}{$\blacktriangledown$ \emph{Open-source General LLMs}} \\
DeepSeek-V4-flash & 28.33 & 36.67 & 25.00 & 45.00 & 25.00 & \textbf{38.33} & 18.33 & 18.33 & 91.67 & 41.67 & \textbf{26.67} & 35.91 \\
KiMi-2.6 & 50.00 & 56.67 & 26.67 & \textbf{73.33} & 58.33 & 36.67 & 16.67 & 15.00 & \textbf{93.33} & \textbf{91.67} & 21.67 & 49.09 \\
Gemma3-27B-it & 22.33 & 27.21 & 25.00 & 33.17 & 21.67 & 31.67 & 20.50 & 16.00 & 77.83 & 39.67 & 24.50 & 30.87 \\
Gemma4-12B-it & 45.00 & 39.50 & 21.17 & 34.33 & 17.67 & 22.17 & 18.50 & 6.17 & 60.33 & 39.67 & 23.50 & 29.82 \\
Qwen3.6-27B & 23.17 & 27.50 & 12.33 & 14.00 & 74.00 & 10.50 & 43.33 & 21.67 & 23.83 & 14.33 & 10.50 & 25.01 \\
Qwen3.5-9B & 27.17 & 25.33 & 26.53 & 23.67 & 30.00 & 18.33 & 45.67 & 27.50 & 27.17 & 20.00 & 21.67 & 26.64 \\
Qwen3.5-4B & 29.17 & 26.83 & 21.17 & 24.67 & 24.83 & 18.17 & \textbf{46.17} & 24.33 & 22.50 & 22.00 & 23.17 & 25.73 \\
\midrule
\rowcolor{tablerowcolor}
\multicolumn{13}{l}{$\blacktriangledown$ \emph{Time-Series LLMs}} \\
TimeOmni-1-7B & 23.67 & 29.00 & 16.50 & 12.33 & 14.33 & 14.50 & 15.33 & 10.50 & 11.67 & 25.33 & 7.67 & 16.44 \\
TS-Reasoner-7B & 29.67 & 28.33 & 27.00 & 28.33 & 26.33 & 27.50 & 23.33 & 27.00 & 38.33 & 35.00 & 25.00 & 28.71 \\
ITFormer-3B & 22.25 & 22.17 & 21.69 & 26.91 & 22.67 & 22.67 & 31.67 & 23.92 & 24.18 & 20.19 & 21.50 & 23.62 \\
ITFormer-7B & 20.46 & 23.43 & 25.30 & 28.44 & 23.29 & 18.33 & 28.88 & 34.58 & 29.35 & 22.40 & 22.33 & 25.16 \\
ChatTS-8B & 21.83 & 23.00 & 25.33 & 20.17 & 25.83 & 13.00 & 24.67 & 33.83 & 42.67 & 25.67 & 21.50 & 25.23 \\
ChatTS-14B & 25.33 & 25.33 & 23.67 & 26.67 & 39.67 & 17.00 & 20.67 & 32.00 & 31.67 & 34.33 & 22.17 & 27.14 \\
Time-LLM & 24.17 & 24.83 & 28.67 & 24.83 & 23.17 & 24.50 & 24.83 & 22.17 & 24.17 & 25.00 & 24.50 & 24.62 \\
AutoTime & 23.33 & 21.50 & 27.33 & 24.33 & 26.33 & 26.00 & 26.83 & 21.84 & 25.83 & 24.67 & 21.67 & 24.51 \\
\bottomrule
\end{tabular}
}
\end{table*}


\subsection{RQ1: How Well Do Current Models Perform on the Benchmark?}
\label{subsec:overall-performance}
We first evaluate all models under the Full-TS setting, where each question is paired with the complete serialized time series. Table~\ref{tab:model_task_accuracy} reports accuracy on each subtask and the macro-average accuracy across all 11 subtasks. Overall, \ours{} remains challenging for current models. The strongest model achieves 50.15\% accuracy, while many open-source LLMs and time-series LLMs perform much closer to the 25\% random baseline. This suggests that \ours{} cannot be solved by directly applying existing general-purpose LLMs or time-series-oriented models.

Performance also varies substantially across tasks. Some models obtain relatively high accuracy on tasks with stronger textual or clinical prior cues, such as time-series summarization or intervention response. However, they struggle on tasks that require precise temporal grounding, forecasting, and decision-making over sparse and asynchronous clinical trajectories. This uneven performance indicates that current models may exploit task-specific cues, but they do not yet provide reliable evidence-grounded reasoning over irregular time series.

\vspace{0.4em}
\noindent
\fcolorbox{black}{gray!10}{%
  \parbox{\dimexpr\linewidth-2\fboxsep-2\fboxrule\relax}{%
    \textbf{\textit{Finding 1:}} Current models remain far from solving \ours{}. Their strong performance on some individual tasks does not translate into reliable temporal reasoning across the benchmark.
  }%
}



\begin{figure}[t]
  \centering
  \includegraphics[width=\columnwidth]{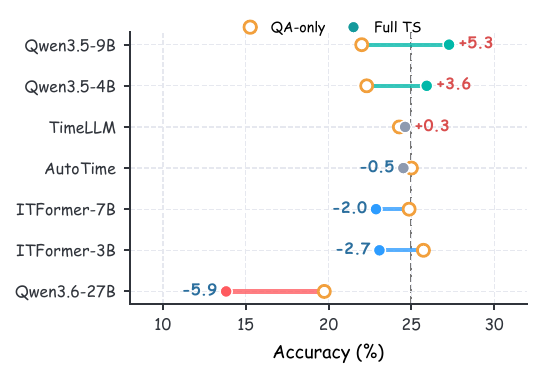}
  \caption{Comparison between QA-only and Full-TS inputs. Each row connects QA-only accuracy to Full-TS accuracy for one model. Rightward movement indicates positive TS Lift (Full-TS minus QA-only), while leftward movement indicates lower accuracy after adding time-series context.}
  \label{fig:ts-lift-dumbbell}
\end{figure}
\subsection{RQ2: Does Full Irregular Context Help?}
\label{subsec:ts-lift}

We next examine whether adding the full irregular time series improves model performance. For each question, we compare two input settings: QA-only, which provides only the question and answer options, and Full-TS, which additionally provides the complete serialized time series. We define TS Lift as the accuracy difference between Full-TS and QA-only.

Figure~\ref{fig:ts-lift-dumbbell} shows that adding the full trajectory produces heterogeneous TS Lift rather than a uniform gain. The Qwen3.5 models improve with Full-TS input, whereas Qwen3.6-27B and ITFormer lose accuracy after the same context is added. Forecasting-oriented TS models remain nearly unchanged. This heterogeneity suggests that irregular clinical context is not automatically usable context. In ICU trajectories, the relevant evidence is often sparse and embedded in long sequences of asynchronous observations, missing values, and non-evidence events. Adding the full time series therefore increases the burden of evidence selection. When models cannot reliably identify the supporting timestamps, the extra context may act as a source of distraction rather than useful information.

\vspace{0.4em}
\noindent
\fcolorbox{black}{gray!10}{%
  \parbox{\dimexpr\linewidth-2\fboxsep-2\fboxrule\relax}{%
    \textbf{\textit{Finding 2:}} Full irregular context is not always helpful. Sparse evidence embedded in irregular time series can introduce context interference and degrade model performance.
  }%
}



\begin{figure*}[t]
  \centering
  \includegraphics[width=\textwidth]{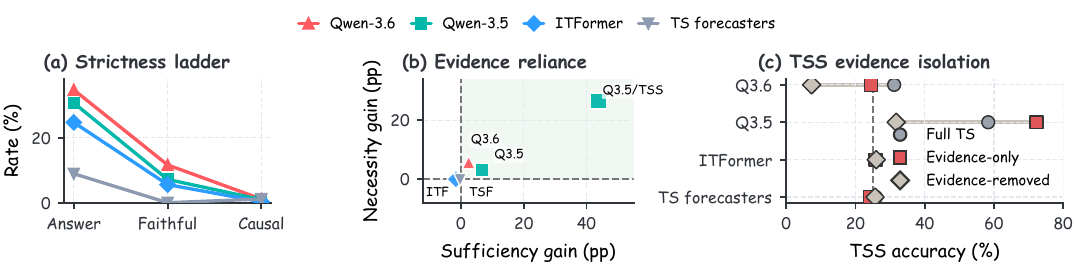}
  \caption{Evidence-use diagnostics across model families. (a) Strictness ladder from answer accuracy to evidence-faithful accuracy and causal sensitivity. (b) Sufficiency (Evidence-only minus QA-only) and necessity (Full-TS minus Evidence-removed), Q3.5/TSS denotes Qwen-3.5 on the TSS task. (c) TSS accuracy under Full-TS, Evidence-only, and Evidence-removed inputs.}
  \label{fig:evidence-failure-board}
\end{figure*}
\subsection{RQ3: Are Correct Answers Grounded in Correct Temporal Evidence?}
\label{subsec:evidence-localization}

We further examine whether models answer correctly for the right reason. For each question, we ask the model to provide both an answer and its supporting evidence. We then compare the predicted evidence with the gold evidence using precision, recall, and F1. We report faithful accuracy, where a prediction is counted as faithful only if the answer is correct and the evidence F1 is at least 0.5. As shown in Figure~\ref{fig:evidence-failure-board}(a), answer accuracy is much higher than faithful accuracy across model families. Even when models choose the correct option, they often fail to identify the timestamped evidence that supports it. This gap shows that answer accuracy alone can overestimate temporal reasoning ability.

We next use evidence ablation to separate two factors: whether the gold evidence is informative, and whether models can find it. \emph{Evidence-only} keeps only the gold supporting evidence, while \emph{Evidence-removed} deletes the gold evidence from the full time series. We define evidence sufficiency and evidence necessity as $\text{Sufficiency} = \text{Acc}_{\text{Evidence-only}} - \text{Acc}_{\text{QA-only}}$ and $\text{Necessity} = \text{Acc}_{\text{Full-TS}} - \text{Acc}_{\text{Evidence-removed}}$. Figure~\ref{fig:evidence-failure-board}(b) shows that most model families have low sufficiency and low necessity. This suggests that they do not consistently benefit from gold evidence when it is provided, nor do they reliably depend on that evidence when it is removed. A clear exception appears in TSS task. As shown in Figure~\ref{fig:evidence-failure-board}(c), Qwen3.5 improves substantially under Evidence-only input and drops when the same evidence is removed. This indicates that the annotated evidence can be useful, but current models often fail to locate and use it within the full sparse and asynchronous record.

\vspace{0.4em}
\noindent
\fcolorbox{black}{gray!10}{%
  \parbox{\dimexpr\linewidth-2\fboxsep-2\fboxrule\relax}{%
    \textbf{\textit{Finding 3:}} Correct answers often do not imply faithful temporal reasoning. Even when annotated evidence is informative in specific tasks, current models do not reliably find and use it in full irregular time series.
  }%
}

\subsection{RQ4: Are Answers Causally Bound to Temporal Evidence?}
\label{subsec:counterfactual}

We next test whether model predictions are causally bound to the temporal evidence that determine the answer. We construct counterfactual pairs for deterministic understanding and reasoning subtasks. In causal edits, we modify the evidence so that the correct answer should change. In irrelevant edits, we modify non-causal observations so that the answer should remain unchanged. We then measure causal flip rate and irrelevant-edit invariance.

\begin{figure}[t]
  \centering
  \includegraphics[width=\columnwidth]{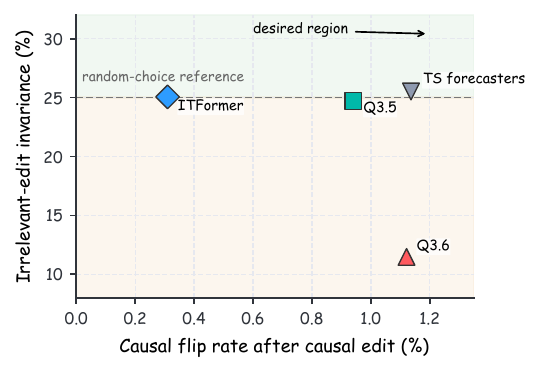}
  \caption{Counterfactual evidence-edit evaluation. Each marker shows a model-family mean. The x-axis reports causal flip rate after causal evidence edits, and the y-axis reports irrelevant-edit invariance.}
  \label{fig:counterfactual-binding}
\end{figure}

Figure~\ref{fig:counterfactual-binding} shows that all model families have causal flip rates below 1.2\%, even when the causal evidence is edited so that the correct answer should change. The mean causal flip rates are 0.94\% for Qwen-3.5, 1.12\% for Qwen-3.6, 0.31\% for ITFormer, and 1.14\% for TS forecasters. Irrelevant-edit invariance is also weak: Qwen-3.5, ITFormer, and TS forecasters stay close to the 25\% random-choice reference, while Qwen-3.6 drops to 11.5\%. Thus, models are neither sufficiently sensitive to causal evidence nor reliably stable under irrelevant changes. The results indicate that models are not reliably binding predictions to patient-specific temporal evidence. Instead, models may rely on clinical priors, option patterns, or global context statistics. This failure is especially important for irregular clinical time series, where the answer may depend on a small number of sparse measurements, short temporal windows, or intervention events at precise timestamps. We conduct this analysis on deterministic understanding and reasoning subtasks, where label regeneration after evidence edits is reliable. Forecasting and decision-making tasks are excluded because their counterfactual labels are not yet safely regenerated.

\vspace{0.4em}
\noindent
\fcolorbox{black}{gray!10}{%
  \parbox{\dimexpr\linewidth-2\fboxsep-2\fboxrule\relax}{%
    \textbf{\textit{Finding 4:}} Current models lack causal sensitivity to temporal evidence. Their answers often fail to change when causal evidence is edited, suggesting that many answers are not truly grounded in patient-specific temporal observations.
  }%
}
\subsection{RQ5: Which Irregularity Mechanisms Expose Hidden Failures?}
\label{subsec:irregularity-stress}

We next examine whether models remain reliable under different forms of temporal irregularity. We stress-test the same Full-TS examples using five perturbations: evidence-preserving missingness, long non-evidence gap injection, timestamp removal, timestamp jitter on non-evidence observations, and observation-order shuffling while keeping timestamps unchanged. For each model--task pair, we report the accuracy change relative to the original Full-TS setting, defined as $\text{Acc}_{\text{perturbed}} - \text{Acc}_{\text{original}}$. Negative values indicate performance degradation, while positive values indicate higher accuracy after perturbation.

\begin{figure*}[t]
  \centering
  \includegraphics[width=\textwidth]{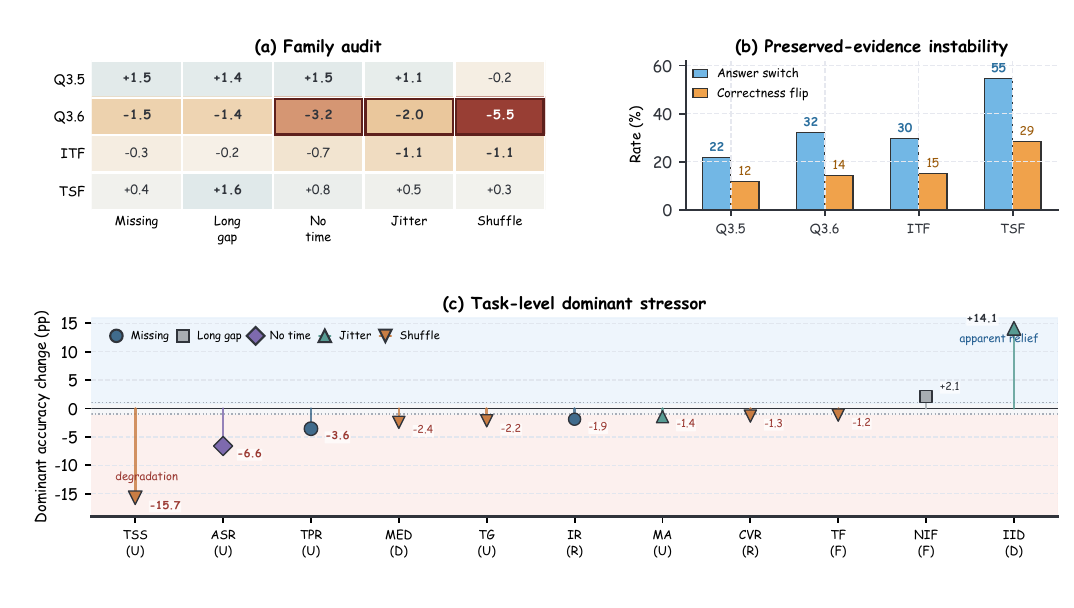}
  \caption{Irregularity stress-test summary. (a) Model-family mean accuracy change under each perturbation, computed as perturbed accuracy minus original accuracy; negative red values denote lower perturbed accuracy, positive blue values denote higher perturbed accuracy, and values near zero denote little mean change. (b) Answer-switch and correctness-flip rates under preserved-evidence missingness. (c) Task-level dominant stress response. Marker type denotes the perturbation with the strongest signed response for each task.}
  \label{fig:irregularity-stress}
\end{figure*}

Figure~\ref{fig:irregularity-stress}(a) shows clear differences across model families. Qwen3.6 is the most sensitive to temporal perturbations: observation-order shuffling decreases accuracy by 5.5\%, timestamp removal by 3.2\%, and timestamp jitter by 2.0\%. ITFormer is more stable on average, but still loses about 1\% point under order shuffling and timestamp jitter. In contrast, Qwen3.5 and TS forecasters sometimes improve under evidence-preserving missingness or long-gap injection. Because the supporting evidence is preserved in these settings, such gains likely reflect shortcut relief rather than genuine robustness. Figure~\ref{fig:irregularity-stress}(c) further shows that different tasks fail under different irregularity mechanisms. Observation-order shuffling is the strongest stressor for TSS, causing a 15.7\% drop. Timestamp removal mainly hurts ASR, with a 6.6\% drop, suggesting that anchor-time reasoning depends strongly on explicit time information. Evidence-preserving missingness most clearly affects TPR, which drops by 3.6\%. In contrast, NIF and IID show positive changes under some perturbations, which we treat as apparent relief rather than evidence of robustness. Finally, Figure~\ref{fig:irregularity-stress}(b) shows that mean accuracy can hide large instability. Under preserved-evidence missingness, TS forecasters switch answers in 54.6\% of grouped examples and flip correctness in 28.5\%. Qwen3.6 also switches answers in 32.2\% of cases. Thus, even when the gold evidence remains available and average accuracy changes only slightly, model predictions can still be unstable.

\vspace{0.4em}
\noindent
\fcolorbox{black}{gray!10}{%
  \parbox{\dimexpr\linewidth-2\fboxsep-2\fboxrule\relax}{%
    \textbf{\textit{Finding 5:}} Temporal-order perturbation dominates many task failures, timestamp corruption selectively breaks anchor-time reasoning, and preserved-evidence missingness exposes hidden answer instability.
  }%
}

\subsection{RQ6: What Is the Accuracy--Latency Trade-off of Text and Time-Series Inputs?}
\label{subsec:input-efficiency}

Finally, we ask whether serialized text inputs and native time-series inputs lead to different efficiency profiles.  We profile 45,725 predictions from two evaluation groups. The text-input group contains 19,730 predictions from Qwen3.5-4B, Qwen3.5-9B, and Qwen3.6-27B, where irregular trajectories are serialized into text. The time-series-input group contains 25,995 predictions from ITFormer-3B, ITFormer-7B, TimeLLM, and AutoTime, which process time-series inputs under their adapted model interfaces. Since these models are evaluated with different inference backends, we report latency as observed end-to-end wall-clock time in our evaluation pipeline, rather than as hardware-normalized model speed.

\begin{figure}[t]
  \centering
  \includegraphics[width=\columnwidth]{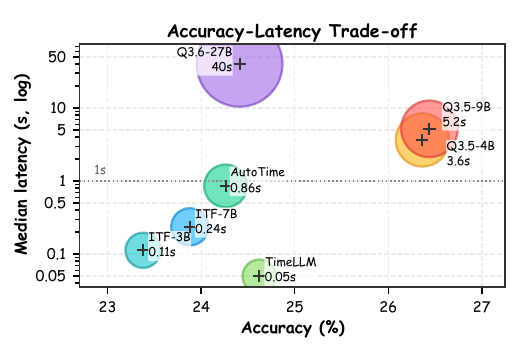}
  \caption{Efficiency analysis.}
  \label{fig:input-efficiency-tradeoff}
\end{figure}

Figure~\ref{fig:input-efficiency-tradeoff} shows a clear accuracy--latency trade-off. Text-input models achieve 25.74\% aggregate accuracy, which is only 1.71 percentage points higher than the 24.03\% accuracy of time-series-input models. However, their median latency is 7.64 seconds per sample, compared with 0.186 seconds for time-series-input models. This corresponds to a 41$\times$ median latency increase for serialized text inputs. The same pattern appears at the model level. Qwen3.5-4B and Qwen3.5-9B reach 26.36--26.44\% accuracy, with median latency between 3.65 and 5.18 seconds. Qwen3.6-27B is much slower, with 40.33 seconds median latency, but does not provide a clear accuracy gain. In contrast, native time-series-input models are much faster. TimeLLM reaches 24.62\% accuracy with 0.050 seconds median latency, while ITFormer variants remain below 0.24 seconds per sample, though their accuracy stays close to the random baseline.

\vspace{0.4em}
\noindent
\fcolorbox{black}{gray!10}{%
  \parbox{\dimexpr\linewidth-2\fboxsep-2\fboxrule\relax}{%
    \textbf{\textit{Finding 6:}} Text serialization provides only a small accuracy gain at a large latency cost. Practical systems require stronger evidence selection or native time-series modeling, rather than simply longer serialized prompts.
  }%
}



\section{Conclusion}
In this work, we introduce \ours{}, a benchmark for irregular clinical time series QA, constructed from de-identified ICU records across four capability dimensions and 11 tasks. Each question is paired with explicit temporal evidence and task-specific answer rules, enabling evaluation not only of answer accuracy but also of evidence use, faithfulness, and causal sensitivity. Through systematic experiments, we show that both general-purpose LLMs and time-series-oriented LLMs remain limited in their ability to perform reliable clinical time-series reasoning. By providing evidence-auditable tasks and diagnostic evaluations, \ours{} offers a focused testbed for developing models that can reason more accurately, faithfully, and efficiently over sparse, asynchronous, and clinically grounded time-series data.

\bibliographystyle{ACM-Reference-Format}
\bibliography{main}

\clearpage
\appendix
\section{Appendix}

\subsection{Ethics and Fairness}
CLIR-Bench is constructed from the de-identified MIMIC-IV ICU database and does not contain directly identifiable patient information. We follow the privacy-preserving principles of the original dataset and use the data solely for research purposes. Nevertheless, clinical datasets may contain inherent biases related to population characteristics, healthcare practices, and data collection processes, which may affect model evaluation and generalization. Therefore, models developed or evaluated on CLIR-Bench should not be directly used for clinical decision-making without further validation and expert oversight. To promote responsible evaluation, CLIR-Bench provides explicit temporal evidence and answer derivation rules, enabling analysis of model faithfulness and reducing the risk of relying on spurious correlations.

\subsection{Data Fields}
\label{apx:data fields}
\ours{} contains routinely recorded ICU vital signs, laboratory measurements, and urine output. The core clinical variables are summarized in Table~\ref{tab:clirbench_variables}. In addition to these variables, \textbf{rel\_hours} records the elapsed time from the beginning of the ICU stay. For each clinical variable listed in the table, a corresponding \textbf{mask} column indicates whether the variable is observed at the current timestamp, while a corresponding \textbf{delta} column records the elapsed time since its previous observation. These auxiliary fields characterize the naturally sparse and asynchronous observation pattern of ICU time-series data. When available, \textbf{intervention\_text} provides textual descriptions of concurrent clinical interventions or events.

\begin{table}[h]
\centering
\small
\begin{tabular}{cc}
\toprule
\textbf{Variable} & \textbf{Meaning} \\
\midrule
$\mathrm{heart\_rate}$ & Heart rate \\
$\mathrm{map}$ & Mean arterial pressure \\
$\mathrm{systolic\_bp}$ & Systolic blood pressure \\
$\mathrm{diastolic\_bp}$ & Diastolic blood pressure \\
$\mathrm{spo2}$ & Peripheral oxygen saturation \\
$\mathrm{resp\_rate}$ & Respiratory rate \\
$\mathrm{temperature}$ & Body temperature \\
$\mathrm{lactate}$ & Blood lactate level \\
$\mathrm{creatinine}$ & Serum creatinine level \\
$\mathrm{wbc}$ & White blood cell count \\
$\mathrm{urine\_output}$ & Urine output \\
\bottomrule
\end{tabular}
\caption{Clinical variables used in the \ours{}.}
\label{tab:clirbench_variables}
\end{table}

\subsection{Construction and Audit Details}
\label{apx:construction_audit}
\paragraph{Candidate generation and evidence grounding.}
We first serialize the irregular patient trajectory using the timestamped fields described in Appendix~\ref{apx:data fields}. Each candidate question is instantiated from a subtask-specific schema, which specifies the queried variable or variables, the model-visible time window, the reasoning goal, the answer format, and the evidence rule. Candidate evidence is selected from timestamped observations, intervention records, or task-specific temporal windows. For forecasting and decision-making questions, we use a cutoff time to separate the model-visible history from the future label-generation window, ensuring that future observations used to determine the answer are not leaked into the input context.

\paragraph{QA construction and automatic filtering.}
Given a valid candidate, Qwen3.6-27B is used to retrieve candidate temporal evidence from the irregular time-series input and draft a multiple-choice question. The correct answer is determined by the task-specific evidence rule and cross-checked against the original timestamped records, rather than relying on the generated text alone. We remove candidates if the evidence window falls outside the patient trajectory, the queried variable is insufficiently observed for the intended task, the intervention context is missing when required, timestamp ordering is inconsistent, the answer is ambiguous, multiple options are correct, or the four options are duplicated or malformed. For forecasting tasks, we additionally check that the answer is derived only from the future label window while the prompt contains only pre-cutoff history.

\paragraph{Distractor strengthening and rewriting.}
After automatic filtering, answer options are shuffled to balance the answer-letter distribution. Distractors are strengthened using clinically plausible alternatives from the same task family, such as nearby value intervals, adjacent temporal windows, confusable trend patterns, or plausible but unsupported intervention choices. GPT-5.5 is then used to rewrite questions and options into concise natural language. The rewriting step is constrained to preserve the original subtask, evidence rule, option semantics, and gold answer label.

\paragraph{Human verification.}
The final construction stage uses human verification as an audit layer. Each retained QA item is checked for four aspects: task suitability, question and option quality, evidence support, and answer-label consistency. Reviewers verify that the item matches the intended subtask, that the question is understandable and the options are mutually exclusive, that the gold answer is supported by timestamped observations or intervention evidence, and that the final answer label remains correct after option shuffling, distractor strengthening, and rewriting. Items with unresolved ambiguity, insufficient evidence, or inconsistent labels are removed.

\paragraph{Evidence metadata and diagnostic input variants.}
Each QA item is associated with evidence metadata used for audit and diagnostic evaluation, including the task name, source stay identifier, model-visible context window, cutoff time when applicable, evidence variables, evidence time span, supporting observations or intervention events, and the deterministic answer rule. These fields support the diagnostic settings used in the experiments: \textit{Full TS} provides the full serialized irregular trajectory; \textit{QA-only} removes the time-series context; \textit{Evidence-only} keeps only the gold supporting observations or events; and \textit{Evidence-removed} removes the gold evidence while preserving the remaining context.

\subsection{Case Study}
\begin{ReasoningBox}{Temporal Grounding}

\centering
\vspace{-0.6em}
\includegraphics[width=1\linewidth]{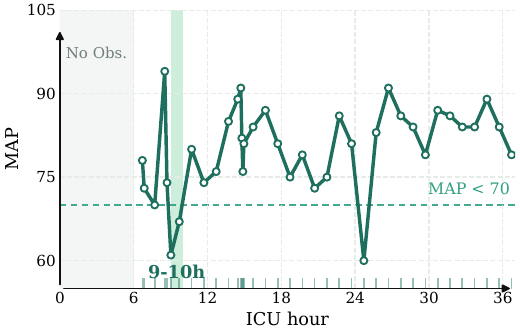}

\vspace{0.5em}
\raggedright
\footnotesize
\textbf{Question:} In the first 37 ICU hours, when is the earliest stretch of observations where mean arterial pressure stays not above 70.00?

\vspace{0.50em}
\textbf{Answer Choices:}\\
\vspace{0.50em}
(A) 9-10h\\
\vspace{0.30em}
(B) 7-8h\\
\vspace{0.30em}
(C) 10-11h\\
\vspace{0.30em}
(D) 8-9h

\vspace{0.30em}
\textbf{Correct Answer:} \textcolor{correctgreen}{(A)}

\end{ReasoningBox}

\begin{ReasoningBox}{Anchor-State Retrieval}
\centering
\vspace{-0.6em}
\includegraphics[width=1\linewidth]{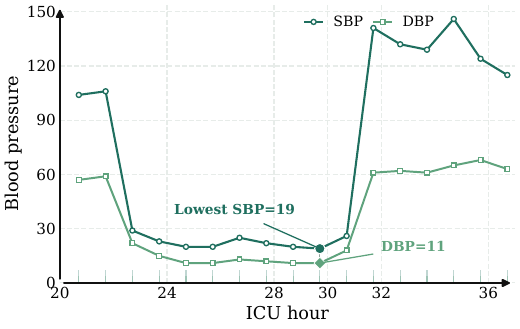}

\vspace{0.5em}
\raggedright
\footnotesize

\textbf{Question:} When SBP is lowest in the episode, which option contains the nearest DBP reading?

\vspace{0.50em}
\textbf{Answer Choices:}\\
\vspace{0.50em}
(A) [11, 15.1)\\
\vspace{0.30em}
(B) [27.2, 35.3)\\
\vspace{0.30em}
(C) [83.9, 92]\\
\vspace{0.30em}
(D) [59.6, 67.7)

\vspace{0.30em}
\textbf{Correct Answer:} \textcolor{correctgreen}{(A)}

\end{ReasoningBox}
\begin{ReasoningBox}{Trend Pattern Recognition}

\centering
\vspace{-0.6em}
\includegraphics[width=1\linewidth]{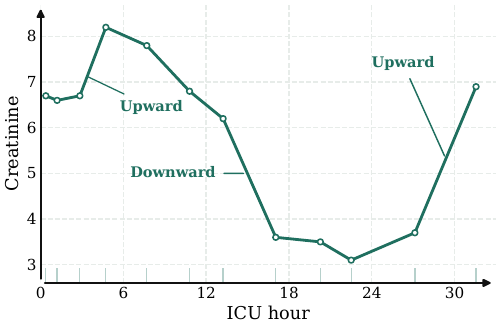}

\vspace{0.5em}
\raggedright
\footnotesize
\textbf{Question:} Which option best summarizes how creatinine changes over time in this ICU series?

\vspace{0.50em}
\textbf{Answer Choices:}\\
\vspace{0.50em}
(A) stable\\
\vspace{0.30em}
(B) upward then downward then upward\\
\vspace{0.30em}
(C) upward then stable then upward\\
\vspace{0.30em}
(D) upward then downward then stable then upward

\vspace{0.30em}
\textbf{Correct Answer:} \textcolor{correctgreen}{(B)}

\end{ReasoningBox}

\begin{ReasoningBox}{Missingness Awareness}

\centering
\vspace{-0.6em}
\includegraphics[width=1\linewidth]{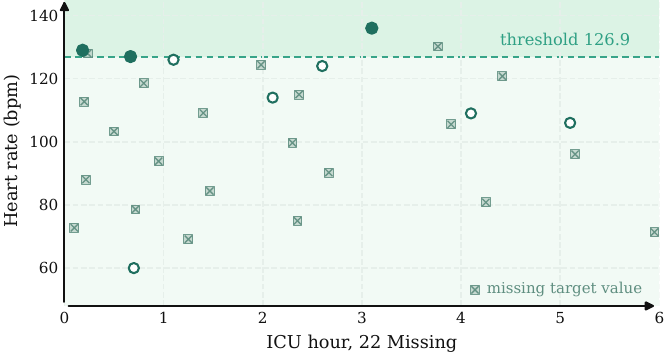}

\vspace{0.5em}
\raggedright
\footnotesize
\textbf{Question:} In the 0:00-6:00 window, can we tell from the available observations whether heart rate was above 126.9?

\vspace{0.50em}
\textbf{Answer Choices:}\\
\vspace{0.50em}
(A) Can determine: the available observations include 2 usable heart rate measurements above the threshold out of 9 usable measurements; 22 heart rate measurements were missing.\\
\vspace{0.30em}
(B) Can determine: the available observations include 4 usable heart rate measurements above the threshold out of 9 usable measurements; 20 heart rate measurements were missing.\\
\vspace{0.30em}
(C) Can determine: the available observations include 5 usable heart rate measurements above the threshold out of 9 usable measurements; 21 heart rate measurements were missing.\\
\vspace{0.30em}
(D) Can determine: the available observations include 3 usable heart rate measurements above the threshold out of 9 usable measurements; 22 heart rate measurements were missing.

\vspace{0.30em}
\textbf{Correct Answer:} \textcolor{correctgreen}{(D)}

\end{ReasoningBox}

\begin{ReasoningBox}{Clinical Time-Series Summarization}

\centering
\vspace{-0.6em}
\includegraphics[width=1\linewidth]{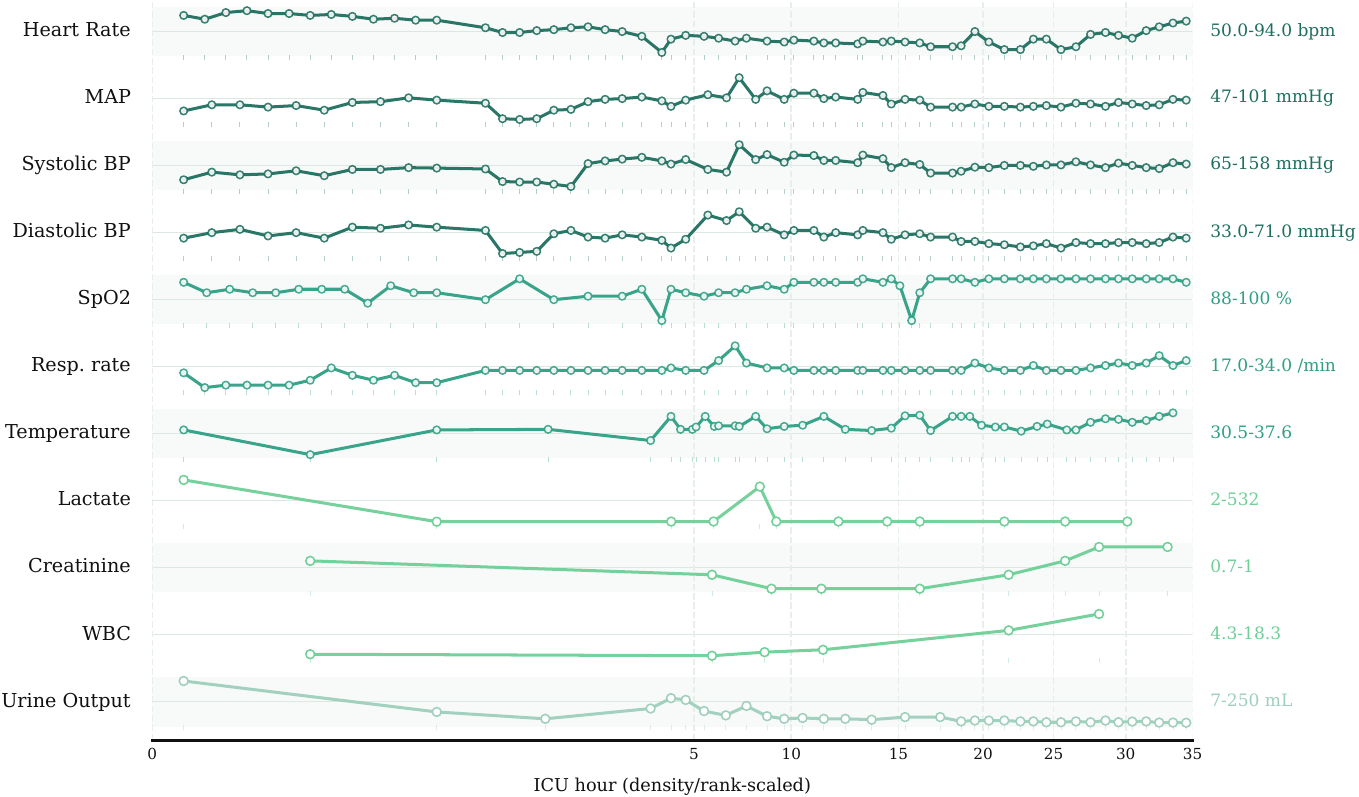}

\vspace{0.5em}
\raggedright
\footnotesize
\textbf{Question:} Which option best summarizes the hemodynamic, respiratory, lab, and urine-output pattern?

\vspace{0.50em}
\textbf{Answer Choices:}\\
\vspace{0.50em}
(A) In broad terms, the main signal is labile hemodynamics with a brief oxygenation drop: systolic pressure fell from 83 to 51, MAP fell from 61 to 37, and heart rate fell from 86 to 47. SpO2 started at 96, ended at 100, and fell as low as 91 around hour 2.3; respiratory rate ranged 18-36. For labs and output, white count rose from 4.3 to 17.5, creatinine was roughly stable (1.0 to 1), lactate fell from 552 to 3.9, and urine output fell from 230 to 8.\\
\vspace{0.30em}
(B) In broad terms, the main signal is labile hemodynamics with a brief oxygenation drop: systolic pressure fell from 78 to 56, MAP fell from 56 to 32, and heart rate fell from 91 to 42. SpO2 started at 100, ended at 97, and fell as low as 86 around hour 2.6; respiratory rate ranged 16-36. For labs and output, white count rose from 5.2 to 17.2, creatinine rose from 0.8 to 1.1, lactate fell from 532.2 to 3.0, and urine output fell from 265 to 7.\\
\vspace{0.30em}
(C) In broad terms, the main signal is labile hemodynamics with a brief oxygenation drop: systolic pressure fell from 80 to 54, MAP fell from 58 to 34, and heart rate fell from 89 to 44. SpO2 started at 99, ended at 99, and fell as low as 88 around hour 2.5; respiratory rate ranged 17-34. For labs and output, white count rose from 4.8 to 18.3, creatinine was roughly stable (0.9 to 1), lactate fell from 532 to 3.4, and urine output fell from 250 to 7.\\
\vspace{0.30em}
(D) In broad terms, the main signal is labile hemodynamics with a brief oxygenation drop: systolic pressure fell from 76 to 50, MAP fell from 54 to 37, and heart rate fell from 94 to 40. SpO2 started at 100, ended at 94, and fell as low as 87 around hour 2.8; respiratory rate ranged 15-38. For labs and output, white count rose from 5.5 to 16.1, creatinine was roughly stable (0.8 to 1), lactate fell from 502 to 2.6, and urine output fell from 280 to 6.

\vspace{0.30em}
\textbf{Correct Answer:} \textcolor{correctgreen}{(C)}

\end{ReasoningBox}

\begin{ReasoningBox}{Asynchronous Cross-Variable Reasoning}
\centering
\vspace{-0.6em}
\includegraphics[width=1\linewidth]{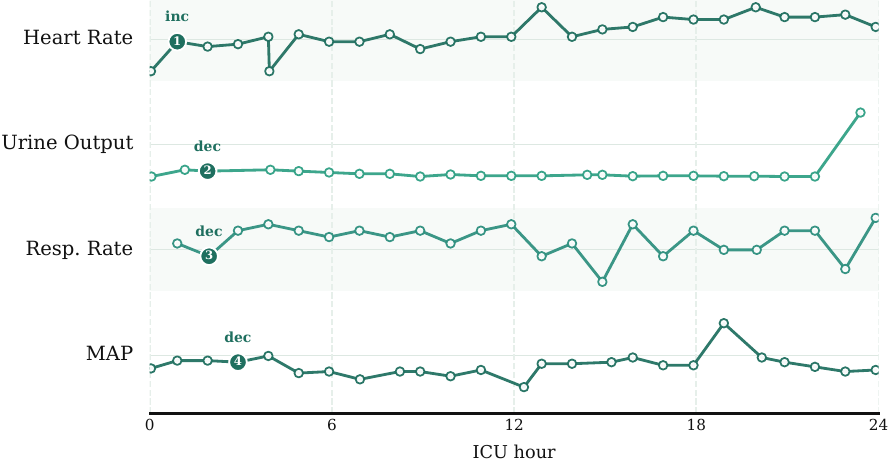}

\vspace{0.5em}
\raggedright
\footnotesize

\textbf{Question:} During the first 24 hours, which answer correctly orders the first observed changes involving heart rate, urine output, MAP, and respiratory rate?

\vspace{0.50em}
\textbf{Answer Choices:}\\
\vspace{0.50em}
(A) heart rate increase occurs first, followed by respiratory rate decrease, then urine output decrease, and finally MAP decrease.\\
\vspace{0.30em}
(B) heart rate increase occurs first, followed by urine output decrease, then respiratory rate decrease, and finally MAP decrease.\\
\vspace{0.30em}
(C) urine output decrease occurs first, followed by heart rate increase, then respiratory rate decrease, and finally MAP decrease.\\
\vspace{0.30em}
(D) heart rate increase occurs first, followed by urine output decrease, then MAP decrease, and finally respiratory rate decrease.

\vspace{0.30em}
\textbf{Correct Answer:} \textcolor{correctgreen}{(B)}

\end{ReasoningBox}
\begin{ReasoningBox}{Intervention Response}

\centering
\vspace{-0.6em}
\includegraphics[width=1\linewidth]{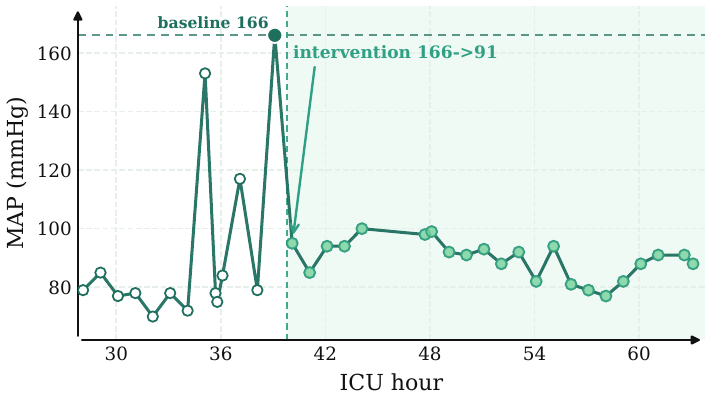}

\vspace{0.5em}
\raggedright
\footnotesize
\textbf{Question:} Once the intervention (inputevents:Dextrose 5\% | 08-Antibiotics (IV) | Fluids/Intake) started at hour 39.8, how did mean arterial pressure move relative to its pre-intervention value?

\vspace{0.50em}
\textbf{Answer Choices:}\\
\vspace{0.50em}
(A) No clear change. The follow-up pattern is not clearly above or below the pre-intervention value.\\
\vspace{0.30em}
(B) Decrease. A brief downward fluctuation is treated as a sustained fall after the intervention.\\
\vspace{0.30em}
(C) Decrease. The follow-up measurements are generally lower than the pre-intervention value.\\
\vspace{0.30em}
(D) Increase. The later measurements are best described as higher overall after the intervention.

\vspace{0.30em}
\textbf{Correct Answer:} \textcolor{correctgreen}{(C)}

\end{ReasoningBox}

\begin{ReasoningBox}{Threshold Forecasting}

\centering
\vspace{-0.6em}
\includegraphics[width=1\linewidth]{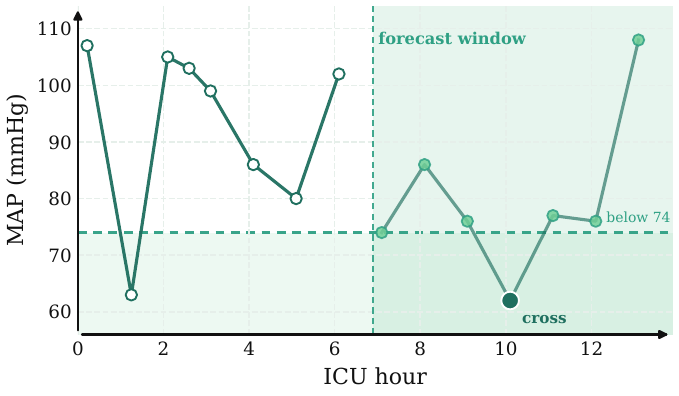}

\vspace{0.5em}
\raggedright
\footnotesize
\textbf{Question:} From the time series available through hour 6.9, will mean arterial pressure be lower than 74 during the next 7 hours?

\vspace{0.50em}
\textbf{Answer Choices:}\\
\vspace{0.50em}
(A) I expect mean arterial pressure to fall below 74 within the next 7 hours; the latest short-term movement before hour 6.9 pointed toward the threshold.\\
\vspace{0.30em}
(B)  I expect mean arterial pressure to stay at or above 74 through the next 7 hours; mean arterial pressure had not been below 74 in the data available before hour 6.9.\\
\vspace{0.30em}
(C) I expect mean arterial pressure to fall below 74 within the next 7 hours; before hour 6.9, arterial pressure had already dropped below the threshold at several time points.\\
\vspace{0.30em}
(D) I would not expect any mean arterial pressure reading to be lower than 74 in the next 7 hours; the recent mean arterial pressure pattern before hour 6.9 was variable rather than steady.

\vspace{0.30em}
\textbf{Correct Answer:} \textcolor{correctgreen}{(C)}

\end{ReasoningBox}

\begin{ReasoningBox}{Next-Value Interval Forecasting}

\centering
\vspace{-0.6em}
\includegraphics[width=1\linewidth]{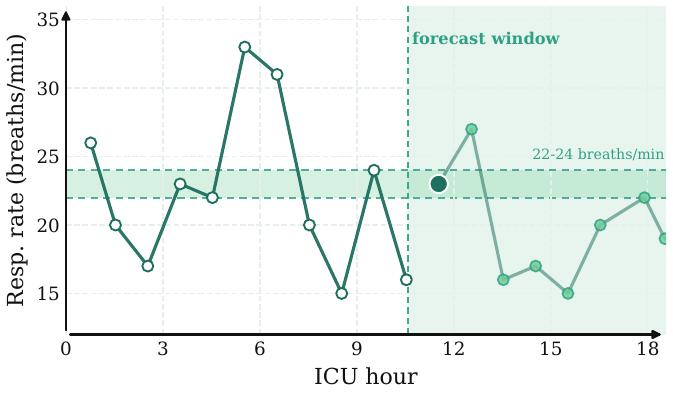}

\vspace{0.5em}
\raggedright
\footnotesize
\textbf{Question:} Given the measurements through ICU hour 10.57, where will the next respiratory rate reading fall in the next 8 hours?

\vspace{0.50em}
\textbf{Answer Choices:}\\
\vspace{0.50em}
(A) 12-14 breaths/min\\
\vspace{0.30em}
(B) 14-16 breaths/min\\
\vspace{0.30em}
(C) 20-22 breaths/min\\
\vspace{0.30em}
(D) 22-24 breaths/min

\vspace{0.30em}
\textbf{Correct Answer:} \textcolor{correctgreen}{(D)}

\end{ReasoningBox}

\begin{ReasoningBox}{Immediate Intervention Decision}

\centering
\vspace{-0.6em}
\includegraphics[width=1\linewidth]{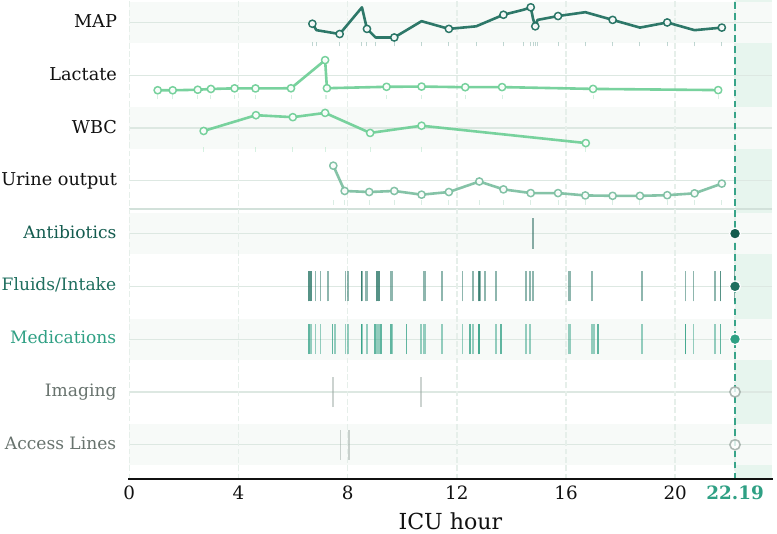}

\vspace{0.5em}
\raggedright
\footnotesize
\textbf{Question:} Based on observations up to hour 22.19, which intervention should be next?

\vspace{0.50em}
\textbf{Answer Choices:}\\
\vspace{0.50em}
(A) Antibiotics; Fluids/Intake\\
\vspace{0.30em}
(B) Antibiotics; Fluids/Intake; Medications\\
\vspace{0.30em}
(C) 5-Imaging; Antibiotics\\
\vspace{0.30em}
(D) Access Lines - Peripheral; Antibiotics; Fluids/Intake; Medications

\vspace{0.30em}
\textbf{Correct Answer:} \textcolor{correctgreen}{(B)}

\end{ReasoningBox}

\begin{ReasoningBox}{Monitoring / Escalation Decision }
\centering
\vspace{-0.6em}
\includegraphics[width=1\linewidth]{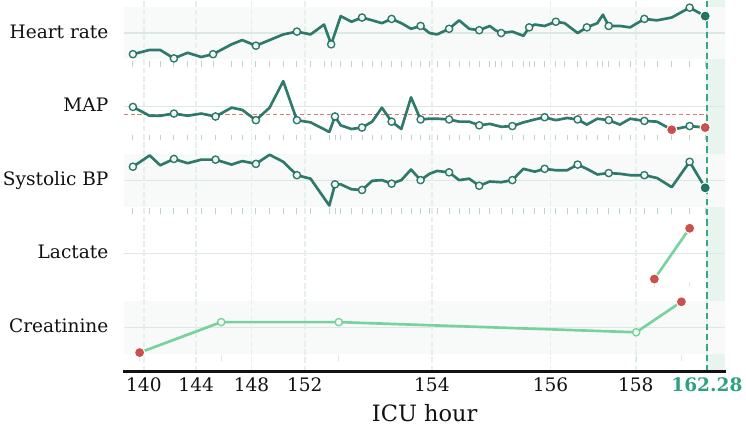}

\vspace{0.5em}
\raggedright
\footnotesize

\textbf{Question:} Based on the observed ICU data so far, choose the next monitoring step: escalate monitoring or continue standard monitoring.

\vspace{0.50em}
\textbf{Answer Choices:}\\
\vspace{0.50em}
(A) Continue standard monitoring for now: Routine bedside checks are reasonable for the next interval. The abnormal reading can be watched at the usual cadence while the bedside values remain steady. Lactate ended at 12.2 mmol/L after dropping as low as 9.1 mmol/L; MAP recovered to 47 mmHg after a recent low of 44 mmHg below 65 mmHg; the current values include MAP 47 mmHg and heart rate 116 bpm.\\
\vspace{0.30em}
(B) Continue standard monitoring for now: The standard monitoring level is reasonable here. The last values do not show a pattern that calls for tighter checks. Creatinine ended at 5.5 mg/dL after dropping as low as 5 mg/dL; lactate ended at 12.2 mmol/L after dropping as low as 9.1 mmol/L; the most recent entries show MAP 47 mmHg and systolic BP 90 mmHg.\\
\vspace{0.30em}
(C) Escalate monitoring now: More frequent checks are reasonable for now. Closer checks help confirm that the apparent recovery holds over the next interval. Lactate ended at 12.2 mmol/L after dropping as low as 9.1 mmol/L; MAP recovered to 47 mmHg after a recent low of 44 mmHg below 65 mmHg; the latest bedside set shows MAP 47 mmHg and heart rate 116 bpm.\\
\vspace{0.30em}
(D)  Escalate monitoring now: Closer monitoring is reasonable for the next interval. The concern is that the latest numbers look better but still follow a recent threshold breach. Creatinine ended at 5.5 mg/dL after dropping as low as 5 mg/dL; lactate ended at 12.2 mmol/L after dropping as low as 9.1 mmol/L; the last snapshot shows MAP 47 mmHg with systolic BP 90 mmHg.

\vspace{0.30em}
\textbf{Correct Answer:} \textcolor{correctgreen}{(D)}

\end{ReasoningBox}

\end{document}